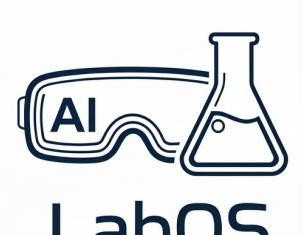

# LabOS: The AI-XR Co-Scientist That Sees and Works With Humans

Le Cong[1,2,**], David Smerkous[1,4*], Xiaotong Wang[1,2,*], Di Yin[1,2,*], Zaixi Zhang[3,*], Ruofan Jin[3], Yinkai Wang[1,2], Michal Gerasimiuk[1,2], Ravi K. Dinesh[1,2], Alex Smerkous[5], Lihan Shi[6], Joy Zheng[6], Ian Lam[6], Xuekun Wu[2,7], Shilong Liu[3], Peishan Li[1,2], Yi Zhu[1,2], Ning Zhao[1,2], Meenal Parakh[8], Simran Serrao[1,2], Imran A. Mohammad[2,9], Chao-Yeh Chen[10], Xiufeng Xie[10], Tiffany Chen[10], David Weinstein[10], Greg Barbone[10], Belgin Caglar[10], John B. Sunwoo[2,9], Fuxin Li[4], Jia Deng[8], Joseph C. Wu[2,7], Sanfeng Wu[6], Mengdi Wang[3,**]

**Affiliations:**

[1] Department of Pathology, Department of Genetics, Stanford University School of Medicine, Stanford, CA, USA

[2] Institute for Stem Cell Biology and Regenerative Medicine, Stanford Cancer Institute, Stanford University School of Medicine, Stanford, CA, USA

[3] Princeton AI Lab, Department of Electrical & Computer Engineering, Princeton University, Princeton, NJ, USA

[4] School of Electrical Engineering and Computer Science, Oregon State University, Corvallis, OR, USA

[5] Department of Bioengineering, University of Washington, Seattle, WA, USA

[6] Department of Physics, Princeton University, Princeton, NJ, USA.

[7] Division of Cardiology, Department of Medicine, Stanford University School of Medicine, Stanford, CA, USA

[8] Department of Computer Science, Princeton University, Princeton, NJ, USA.

[9] Department of Otolaryngology-Head and Neck Surgery, Stanford University School of Medicine, Stanford, CA, USA

[10] NVIDIA. Santa Clara, CA, USA

[*] Core contributing authors, listed alphabetically with equal contribution.

[**] Corresponding authors: Le Cong (congle@stanford.edu), Mengdi Wang (mengdiw@princeton.edu)

## Abstract

Modern science advances fastest when thought meets action. LabOS represents the first AI co-scientist that unites computational reasoning with physical experimentation through multimodal perception, self-evolving agents, and Extended-Reality(XR)-enabled human-AI collaboration. By connecting multi-model AI agents, smart glasses, and robots, LabOS allows AI to see what scientists see, understand experimental context, and assist in real-time execution. Across applications--from cancer immunotherapy target discovery to stem-cell engineering and material science -- LabOS shows that AI can move beyond computational design to participation, turning the laboratory into an intelligent, collaborative environment where human and machine discovery evolve together.

## Introduction

Science advances on two coupled fronts: computation that proposes and predicts, and experimentation that validates and reveals. Recent AI has transformed the computational front [1,2,3,4]—accelerating simulation, prediction, and design—yet the physical laboratory remains the bottleneck, where perception, coordination, and reproducibility still limit progress, and outcomes hinge on hard-to-transfer or tough-to-reproduce skills. Meanwhile, today's "agentic AI" largely operates in the digital realm—planning experiments and synthesizing tools from text, data, and simulations without perceiving or acting in the dynamic lab. In parallel, robotic automation in laboratories can be powerful but mostly rule-based and bespoke—costly to retarget, challenging to relocate, and brittle to real-world variability.

LabOS addresses this gap through a unified human–AI collaborative intelligence platform that makes laboratories AI-perceivable and AI-operable. It integrates *agentic AI systems* for digital lab reasoning with extended reality(*XR)-enabled, multimodal interfaces* for human-in-the-loop physical lab execution, creating an end-to-end framework that links hypothesis generation, experimental design, physical validation, and automated documentation.

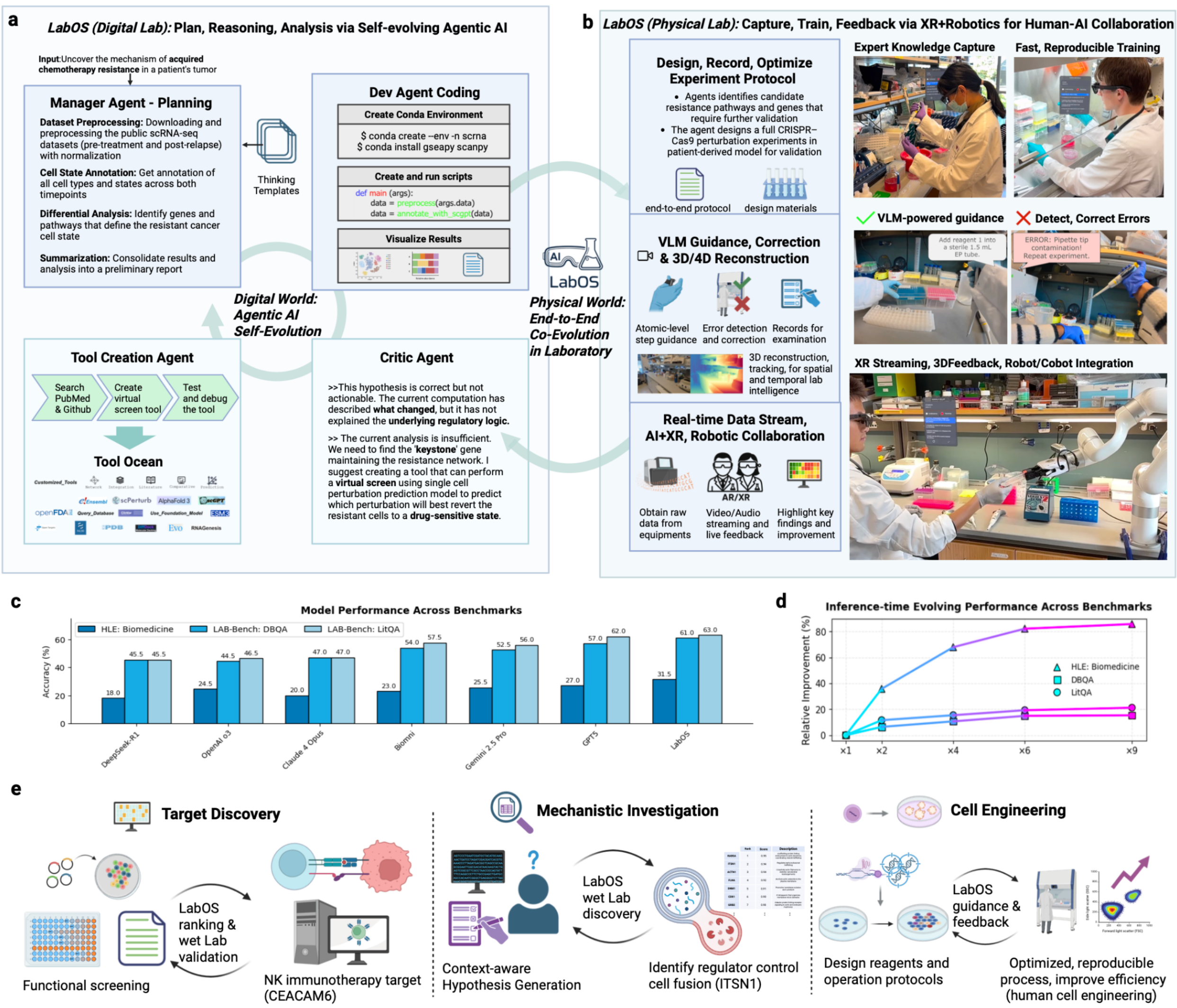

**Figure 1. LabOS: Multi-Modal Human-AI Collaboration in Science Laboratory.** LabOS comprises a self-evolving agentic AI for digital lab tasks and an XR interface for human-in-the-loop physical lab execution, creating an end-to-end system for lab research. **a, Digital (Dry) Lab:** Agentic AI as computational core of LabOS. The system employs a multi-agent framework, including a Manager Agent

for planning, a Dev Agent for coding and execution, and a Critic Agent for evaluation and refinement. The agents continuously improve analytical capabilities. A Tool Creation Agent expands the system's functions by generating new tools from sources like PubMed, which are then stored in a shared Tool Ocean. This component automates complex data analysis. **b, Physical (Wet) Lab:** Capture, guide, live feedback via AR/XR for human-AI collaboration in the physical laboratory. A scientist wearing AR/XR glasses receives real-time guidance from the AI. A specially trained Vision-Language Model (VLM) monitors the procedure, providing on-the-fly error detection and correction to ensure correctness and reproducibility. Finally, the LabOS Robot/Cobot module allows AI agents to plan and guide human scientists to collaborate with robotics, enabling robotic automation of time-consuming steps and seamless human-AI handover and collaboration **c,** Results on leading benchmarks—*HLE: Biomedicine*, *LAB-Bench: DBQA*, and *LAB-Bench: LitQA*—show LabOS outperforming frontier LLMs/agents in biomedical reasoning tasks. **d,** Self-evolving agent's performance scales with inference-time compute. **e, Use Cases.** The first case is drug target discovery: the agentic AI analyzes functional screening data to identify and rank NK cancer immunotherapy targets. Secondly, for mechanistic investigation, LabOS generates testable hypotheses that are then validated by a human scientist to identify a cell fusion regulator. In the physical lab execution for these use cases, LabOS enabled copiloted, reproducible processes for complex experiments like human cell engineering. **All photographs shown are of the authors.**

The LabOS co-scientist adopts a multi-agent reasoning architecture—comprising planning, development, critique, and tool-creation agents—that collectively perform hypothesis generation, experimental design, data analysis, and adaptive improvement. The AI co-scientist is self-improving, continuously expanding its analytical capabilities via a “Tool Ocean” of modules autonomously generated from web search, scientific literature, and data. This self-evolving capacity enables the AI to solve novel research tasks via inference-time scaling.

In this paper, we focus on an end-to-end instantiation of LabOS for the biomedical domain and demonstrate state-of-the-art performance on leading biomedical reasoning benchmarks—including Humanity’s Last Exam (HLE): Biomedicine [5], LAB-Bench: DBQA, and LAB-Bench: LitQA [6]—while closing the loop from digital lab planning to physical lab execution.

On the physical side, LabOS connects AI reasoning directly to the laboratory via AR/XR smart glass interfaces and real-time multimodal sensing. Researchers wearing XR glasses receive adaptive, context-aware guidance from the AI agent—step-by-step instructions, error detection and correction cues, and gesture or voice interactions for sterile workflows.

To enable the AI co-scientist to “*see”* in the lab, we collected >240 egocentric video sessions from researcher-worn cameras/glasses during real experiments, assembling these into the LabSuperVision (LSV) benchmark for evaluating AI models’ lab perception and reasoning capabilities. Because leading AI models struggled on this benchmark, we post-trained a lab-specialized Vision-Language-Model (VLM) using a combination of public experimental videos, in-house recordings, and expert annotations. The resulting model, namely the LabOS VLM, is able to decode visual input from XR glasses and align the visual embedding with a language model to interpret and reason about lab video scenes. The model demonstrates markedly improved visual reasoning capability in scientific settings, enabling LabOS to monitor actions, detect deviations, verify results, and synchronize multimodal data streams with reference protocols—allowing the AI to perceive, understand, and act/co-pilot within real laboratory as an AI co-scientist.

LabOS also supports 3D/4D spatial modeling of laboratory workflows. These digital twins capture spatial and temporal relationships between instruments, samples, and human actions, enabling replay, “what-if” analysis, and simulation-based training. The resulting spatial grounding provides the foundation for safe, reproducible, and transferable laboratory automation.

For real world validation, we demonstrate LabOS's ability in three biomedical research studies: cancer immunology, mechanistic research, and stem cell engineering (**Fig. 1c**). In the first study, we sought to uncover cancer immunotherapy targets that could boost natural killer (NK) cell killing of tumors. We used LabOS to generate hypotheses, perform target identification via multi-step reasoning and analysis, and the AI agent nominated CEACAM6 as a putative target, which were validated in physical lab NK-tumor killing assay. In the second study, we used LabOS to perform mechanistic research, showcasing identification of a gene regulating cell fusion, ITSN1. In the third study, two researchers wore smart glasses during stem-cell engineering projects and interacted with the AI co-scientist. The latest LabOS streams egocentric data to the server and invokes the VLM agent every ~5-10s. In a gene-knockout experiment, LabOS monitored the workflow, provided step-level guidance, and could flag operation deviations—such as when the researcher did not follow sterile techniques, or, wrong reagent incubation time versus gold-standards. In a lentiviral transduction in human iPSCs experiment, LabOS was used to automatically record and digitalize workflows by an experienced researcher, which can be used for coaching a novice researcher to complete the experiment, demonstrating its potential in capturing and transferring advanced human skills.

Together, these features make LabOS a true AI co-scientist—the first multimodal human–AI system to unify digital lab reasoning and physical lab execution in a single, adaptive framework. By giving AI the ability to think with us and work alongside human scientists—seeing what we see, checking what we do, and learning from every run—LabOS turns the lab into a dynamic feedback loop. It moves us toward autonomous, self-improving discovery, where human intuition and machine rigor co-evolve to accelerate breakthroughs.

## Results

### 1. LabOS Overview

LabOS links agentic AI for digital lab reasoning with XR-enabled, human-in-the-loop physical lab execution into an end-to-end workflow from design to validation. A multi-agent core—Manager (planning), Developer (coding/execution), and Critic (evaluation)—plus a Tool-Creation module that auto-extends a shared Tool Ocean nominates targets, plans experiments, runs analyses, and continually improves [7,8] (**Fig. 1a**).

For multimodal human-AI collaboration in the physical lab module, protocols can be generated by or input into LabOS, then streamed to the connected XR glasses. The interface on XR glasses (i) renders stepwise protocol in a Unity/Android application, (ii) verifies physical actions from the first-person video stream by invoking an embedded VLM for visual reasoning, and (iii) returns context-aware feedback in real time (**Fig. 1b**). All streams are time-stamped and logged with metadata for automated documentation. To further enhance the human-AI collaboration in physical lab, we also developed the LabOS robotics module, where agentic AI could control a cobot (e.g. a robotic arm) to automate time-consuming or repetitive steps within a protocol, with seamlessly hand-over between human and robotic system driven by XR-VLM integration and the reasoning/planning capacity of the LabOS agent (**Extended Data Fig. 1**).

### 2. LabOS Supports Self-Evolving AI Agent for Biomedical Reasoning

The digital-lab module of LabOS builds on and expands the STELLA self-evolving agent framework [8], a multi-agent reasoning system for biomedical research that integrates planning, development, critique, and tool creation. **Fig. 1(a)** illustrates the agentic architecture. The Manager/Planner Agent decomposes scientific objectives into structured modules—candidate molecules, reagents, procedures, materials lists, instrument settings, and quality control checkpoints. The Developer Agent executes these steps by generating and running Python code for complex bioinformatics analyses, while the Critic Agent evaluates intermediate results and refines the workflow, forming an iterative reasoning loop.

This AI co-scientist learns and improves from every problem it solves, continuously enhancing its technical abilities by proactively expanding reasoning strategies and creating new tools [7,8]. Two mechanisms drive its continuous self-improvement. First, a Template Library of successful reasoning workflows is dynamically updated, allowing the system to generalize from prior solutions. Second, a Tool Ocean maintains and expands a repository of analytical tools/codes, databases, and APIs. The Tool-Creation Agent autonomously identifies, tests, and integrates new resources as needed. This architecture enables LabOS to plan and solve complex research tasks efficiently while continually enhancing its reasoning and analytical capabilities.

To evaluate the efficacy of LabOS research agent, we test-ran it on a suite of challenging biomedical reasoning benchmarks. Our results show that LabOS consistently establishes a new state of the art, achieving top accuracy scores of approximately 32% on *Humanity's Last Exam: Biomedicine*, 61% on *LAB-Bench: DBQA*, and 65% on *LAB-Bench: LitQA*, outperforming the next-best models by up to 8% (**Fig. 1c**). Crucially, its performance systematically improves with continued use and test-time scaling, providing direct evidence of its self-evolving design (**Fig. 1d**). The self-improving framework enables LabOS to learn and grow like a human scientist, dynamically scaling to meet the ever-expanding challenges of biomedical discovery.

## 3. Training LabOS AI Co-Scientist To See and Reason in Physical Labs

To enable LabOS to see, understand, and reason in a physical lab environment, we sought to post-train a Vision-Language-Model (VLM) on a broad set of lab research videos. A VLM is a multimodal AI model that jointly learns from visual and textual inputs, allowing it to connect what it *sees* with what it *reads* or *describes*. It typically combines a vision encoder that processes images or video frames with a language model that interprets and generates text. By aligning these two modalities in a shared representation space, the VLM can recognize lab scenes, interpret actions, and reason about experimental workflows and outcomes.

### 3.1 LabSuperVision (LSV): Benchmarking AI for Scientific Visual Reasoning

We asked researchers to wear XR glasses or action cameras when they operate in their laboratory research to collect real-world data. Then we processed and annotated the collected data for evaluating AI models in lab research.

We assembled **LabSuperVision (LSV)**, an expert-annotated laboratory video dataset, designed for lab operation video understanding and reasoning. LSV comprises >200 high-quality video sessions (typically 2–10 min; up to 45 min) recorded by 7 researchers across diverse lab scenes—bench, tissue culture room, and instrument bays—capturing realistic operations and movement between spaces. Each session is associated with an expert-generated gold-standard protocol. Then, human scientists performed corresponding experiments wearing cameras or smart glasses, recording all details. For annotation, a team of five human experts annotate each video session with: (1) Step segments with start/stop times aligned to the gold-standard protocol. (2) Error and issue events labeled by type (e.g., sterile breach, step mismatch, timing deviation). (3) Critical parameters, materials and reagents when applicable (**Fig. 3**).

With the LSV benchmark, we evaluate whether leading commercial AI models can interpret and troubleshoot scientific procedures from lab experiment videos. We tested four leading AI models spanning both open and proprietary foundation models: Gemini 2.5 Pro, Cosmos-Reason-1, Qwen 2.5-VL-7B, and GPT-4o (**Fig. 3c**). When prompted with a new video, each model is tasked with one of the two: (1) Protocol alignment, where the model needs to generate a stepwise protocol describing procedural actions and parameters, (2) Issue identification, where the model needs to troubleshoot based on the video and identify any deviation from gold-standard or any handling errors if present. The model outputs are then compared against the ground truth, using two methods: a total of five human experts compared VLM outputs vs. gold-standard protocol and error labeling following a 0-5 scoring (5 being the highest for perfect alignment, detailed rubrics in Supplementary Material); in parallel, we used GPT-5 as a comparator to examine and score outputs using the same rubrics.

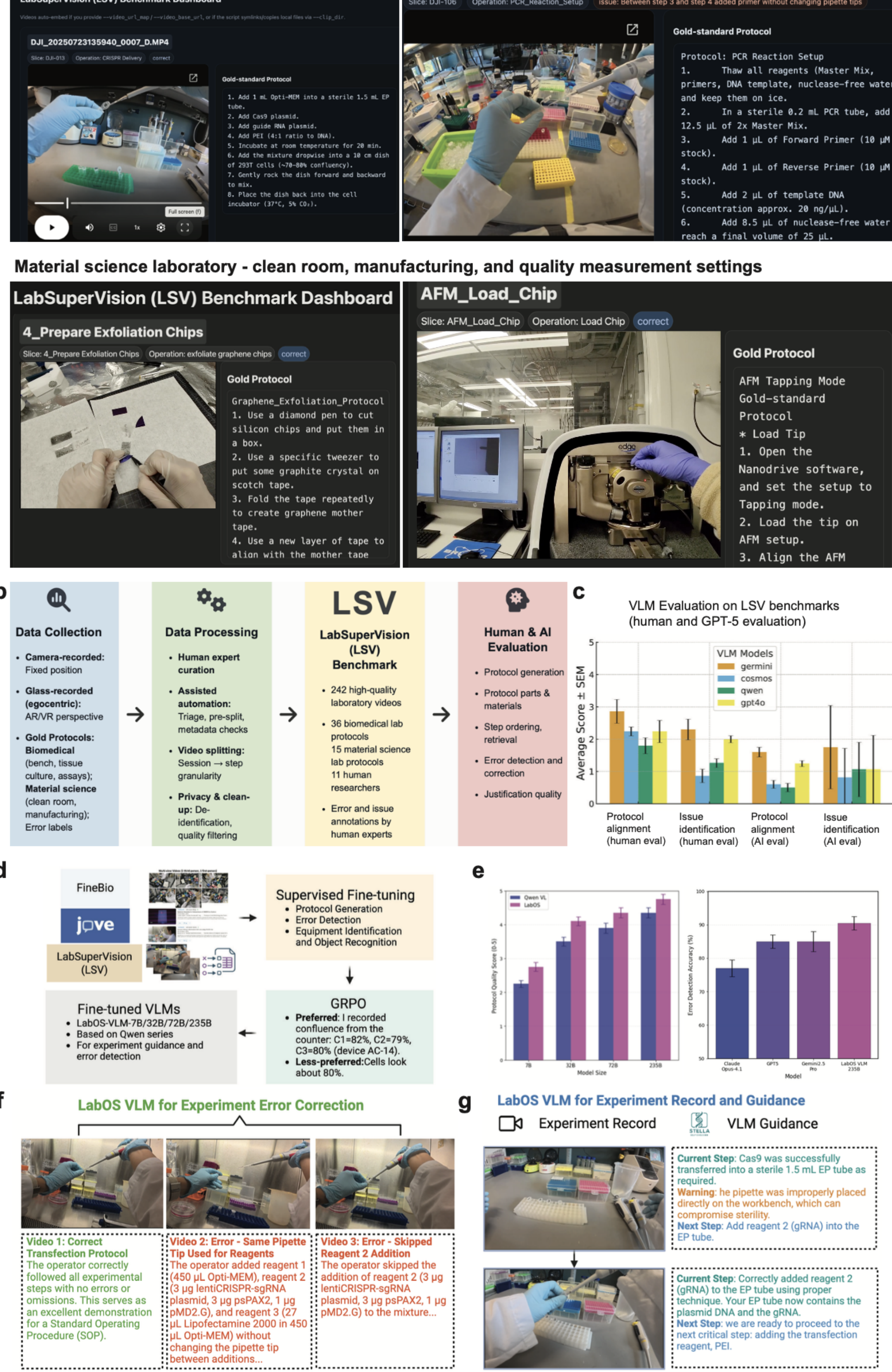

**Figure 2. LabOS-VLM for visual reasoning in laboratories. a,** LabSuperVision (LSV) benchmark dashboard with egocentric and fixed-view lab videos paired to gold-standard protocols, including notes,

issues or errors labeled by human experts. We collected data across multiple laboratory settings spanning biomedical labs (top) and material science labs (bottom). **b,** LSV benchmarking pipeline in four stages: (1) Data collection—multi-modal recordings from diverse facilities via fixed cameras and XR smart glasses; (2) Data processing—expert curation with light automation, aligned to reference protocols and parameters; (3) Dataset assembly—compressed corpus of 200+ experiment-session videos; (4) Evaluation—human and GPT-5 assessments of model performance on LSV. **c,** Benchmarking AI models on LSV via human experts (n=5) and GPT-5. Scores range 0–5 (5 = perfect). **d,** LabOS-VLM post-training using FineBio, JoVE, and LSV datasets via supervised fine-tuning (SFT) and reinforcement learning (GRPO). **e,** LabOS-VLM family outperform baselines on protocol generation quality (left) and error-detection rate (right) on LSV. **f–g,** Real-world testing of LabOS-VLM: (f) error detection/correction in a tissue culture experiment; (g) context-aware AI generated step-by-step instructions grounded in visual understanding.

Our findings show that leading AI models struggle to understand fine-grained laboratory research workflows: the top-performing model, Gemini-2.5Pro, scored only 2.86 out of 5 in protocol alignment, moderately better than open-source NVIDIA Cosmos-1 which scored 2.24; for issue/error identification, leading models like Gemini, GPT4o only managed to score ~2 out of 5 (**Fig. 2c**). This reflects the gap between open-world foundation models' capabilities and what is needed for scientific research in physical labs. We performed similar tests in material science labs (**Fig. 2a**) and reached the same conclusion. Our results surface where today's AI models have some success (protocol alignment) and where they have substantial limitations (error detection/correction).

### 3.2 LabOS-VLM: Vision-Language-Model Trained for Visual Reasoning in Laboratories.

Given the aforementioned gap, we sought to train a vision–language model for visual reasoning in lab environments. For this purpose, we assemble an experimental video collection, including FineBio [16] (expert-annotated physical lab videos), JoVE (standardized procedure videos), and LSV. We split the datasets 80/10/10 for training, validation, and held-out testing.

Using Qwen-VL as the base model, we performed post-training by supervised fine-tuning (SFT) with LoRA on paired video–text examples and then reinforcement finetuning to improve visual reasoning. In the latter reinforcement learning step, we used Group Relative Policy Optimization (GRPO) [9] with LoRA updates, where the model rolled out a group of multiple candidate responses per prompted scenario and received rewards designed to evaluate the candidate outputs. The reward is designed to be rule-based to account for lab procedural accuracy, safety compliance, and experimental detailed level; as well as to account for relative rewards within each group to favor expert-consistent reasoning. By using this SFT->RL pipeline, we adapted the base model and obtained the LabOS-VLM (7B, 32B, 72B, 235B; **Fig. 2e–f**) family.

Across model scales, the LabOS-VLM family consistently outperforms the base model on lab video reasoning tasks (measured via protocol generation quality and error detection accuracy). On the held-out subset of evaluation data, LabOS VLM-235B achieves >90% accuracy in error detection accuracy, outperforming Claude Opus-4.1, GPT-5, and Gemini 2.5 Pro on all evaluated metrics, establishing LabOS VLM as a strong module for multimodal lab intelligence and AI-assisted experimentation (**Fig. 2e**). Our SFT->RL pipeline has proved to enhance the AI model's ability in step-wise visual reasoning, error detection and correction, which is a basis for building AI co-scientist to see and work with researchers real-time in scientific settings.

We next evaluated the fine-tuned LabOS-VLM on egocentric videos from real experiments. First, we supplied a series of videos about delivering CRISPR gene-editors to human cells via transfection. The model successfully distinguished correct vs. incorrect operations, pinpointing the specific errors that the human operator made across 2 distinct issues (**Fig. 2f**). Second, we provided experiment records, from a multi-step setup on preparing Cas9 RNA complex for gene-editing along with a reference protocol, to the LabOS co-scientist. The VLM-powered AI agent recognized each step, generated step-wise guidance, matched actions to the protocol, issued context-aware warnings, and suggested the next action. These results validate LabOS VLM capabilities in authentic physical lab settings.

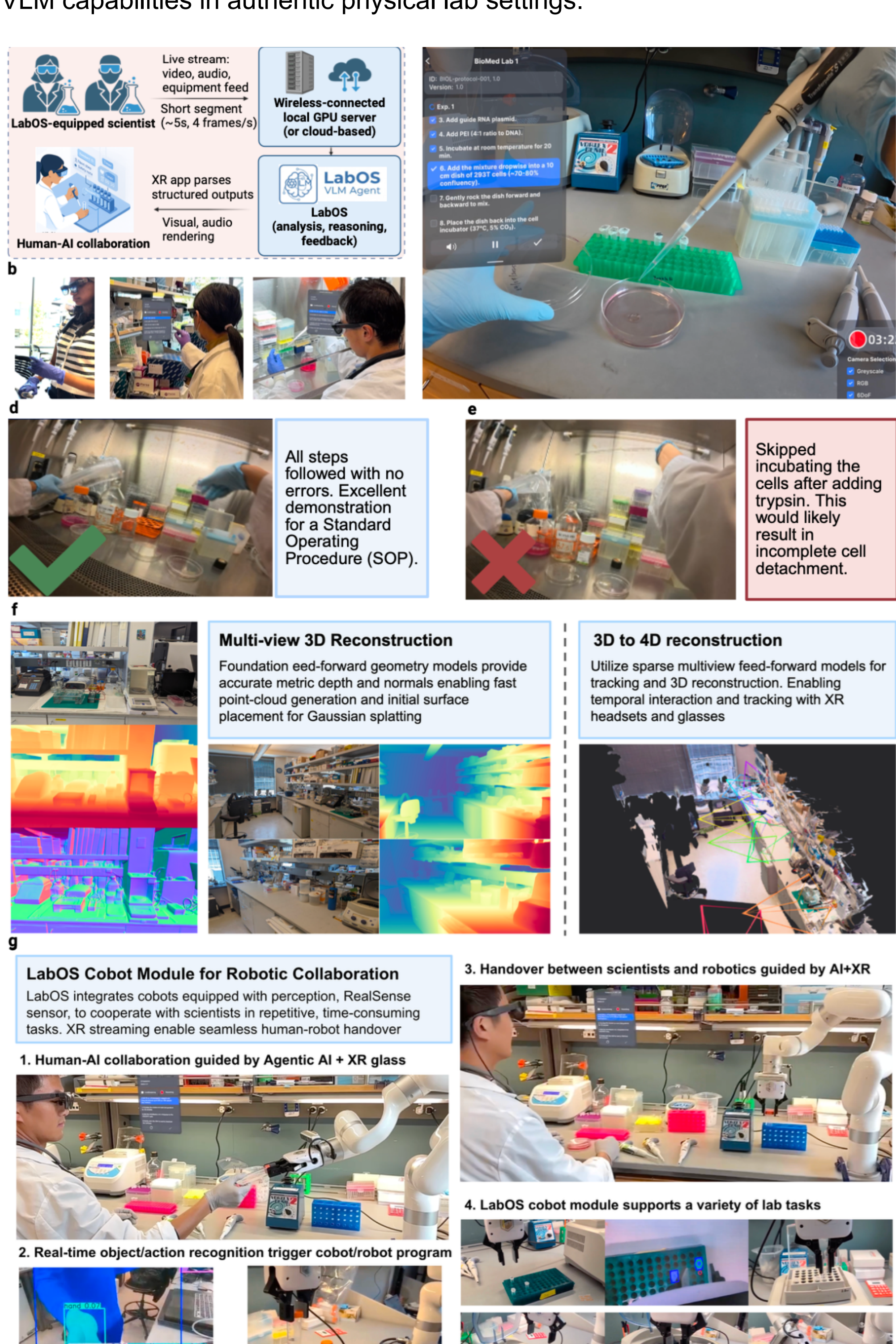

**Figure 3. LabOS on XR glasses enables spatially-grounded human–AI collaboration in physical laboratories. a,** Live streaming from XR glasses to the LabOS VLM server enables real-time scene understanding, feedback, and interactions between human and AI agents. **b,** Deployment of LabOS AI+XR system across lab settings. **c,** Live action feed from the wearer's perspective. **d,** LabOS AI+XR provides guidance and summary on the lab operation. **e,** Detected deviations trigger inline error prompts and suggested corrections, to mitigate human researcher's oversight. **f,** Feed-forward geometry models supporting camera tracking and multi-view, metric-depth 3D reconstruction, enabling object localization and spatio-temporal reasoning for scientific workflows. **g** Proof of concept adaptive cobot module utilizing open vocabulary vision models to interact with humans while leveraging robotics operation to streamline time-consuming steps on existing lab equipment. For this demo we utilized an xArm + gripper equipped with an Intel Realsense camera.**All photographs shown are of the authors.**

## 4. Extended-Reality (XR) Glasses Enable Real-Time Human-AI and Robotic Collaboration

Human–AI interaction in LabOS is mediated through the XR interface. Researcher engages with the AI co-scientist primarily via the LabOS XR glasses, where the glasses live-stream what the human scientist sees and hears while reconstructing the 3D lab environment. Modern XR hardware supports interface rendering, gesture recognition, and running an Unity/Android application. Data from the glasses are streamed to a local GPU server (or the cloud) for real-time agentic inference. The server receives short video segments (e.g., 5–10s), forwards them to the LabOS AI agent for analysis and reasoning (input at 4 frames/s for videos), and returns a structured JSON output to the XR glasses. The XR app parses this JSON message and provides real-time visual and audio feedback to researchers at the bench (**Fig. 3a-e**). We tested both AR/XR glasses and VR/XR headsets, while both supported the above functions, we chose to launch initial deployment of LabOS using AR/XR glasses as they achieved key specifications needed for human-centered intelligent lab: open-peripheral, light-weight hardware (less than 3-ounces/85-grams) allows convenient wearing in labs where goggles are already used; 2+ hours battery life for extended operation (via a wireless neckband or mobile battery); display brightness over 1200+ Nits sufficient for in-door labs; 6DoF and hand gesture support for 3D-aware human-AI interactions. We validated that the LabOS AI-XR solution can support error detection, correction, guidance in various experimental tasks (**Extended Data Fig. 2**).

Through real-time supervision and guidance, the LabOS AI Co-Scientist boosts reproducibility by standardizing experiment capture and logging non-fatal deviations and context variables. Expert-run recordings from XR glasses also serve as instructional material and training data to improve LabOS agents and downstream models.

LabOS also supports 3D modeling of lab environments and workflows, for a "world model" of scientific labs (**Fig. 3f**). This model is generated via videos captured on researcher-worn cameras or glasses — optionally augmented with multiview cameras — using state-of-the-art 3D/4D reconstruction algorithms. In our experiment, we utilize MapAnything [10] for multiview environment + egocentric for tracking, camera positioning, depth maps, and point cloud reconstructions, and incorporate 3D Gaussian splatting. Gaussian splatting models scenes as sets of millions of Gaussian distributions whose parameters are inferred from input 2D images captured in different camera positions and (and time if needed), enabling photorealistic, temporally consistent reconstruction. 4DLangSplat [11] can further help produce a time-aware, semantically indexable 3D environment, to support object-centric tracking. Additionally, to track the complex lab operation and human motion, we show the use of off-the-shelf models for hand tracking and in-hand object reconstruction (**Extended Data Fig. 3, Supp. Video 1**). We employed HAPTIC [17] to track hand motion and MegaPose [18] to obtain the 6-DoF pose of the object (in the demo case, a pipette) using scanned CAD models. Results from HAPTIC and MegaPose may be misaligned due to the lack of a common scale. To align the hand–object motion, we use HORT [19] to estimate the scale of the object motion relative to the hand.

Further, , utilizing the 3D/4D AI perception through camera feeds, we demonstrate a proof-of-concept cobot module for LabOS, as in (**Fig. 3g, Supp. Video 2**). This module enables scientists to run time-consuming and/or repeated portions of the protocol using an adaptive, and spatially aware robotic arm. We demonstrate example protocols loaded in LabOS, such as vortexing, 96-well plate operations, tube handling on incubator/shaker, can be handed off to the cobot to reduce the load on scientists, where agentic and embodied AI with spatial intelligence provide a seamless human-AI-robotics collaboration experience (**Supp. Video 3**).

## 5. LabOS In Action: Human-AI Collaboration for Biomedical Discovery

### 5.1 LabOS for cancer immunotherapy target discovery

The challenge in cancer immunology is to identify regulators of immune evasion that mediate tumor resistance to cytotoxic lymphocytes such as natural killer (NK) cells. Traditional screening approaches are limited by throughput and rely on human expertise for downstream analysis. We prompted LabOS to investigate genes regulating sensitivity and resistance to NK cell-mediated killing of melanoma cancer cells based on a functional screen.

Using a functional CRISPR activation (CRISPRa) screen in A375 melanoma tumor cells, treated with or without primary human NK cells, LabOS AI co-scientist autonomously identified and refined candidate regulators of NK-mediated cytotoxic resistance [12]. Within a lab-in-the-loop pipeline, the LabOS AI agent dynamically re-ranked genes through iterative functional enrichment analysis, revealing CEACAM6 as a top regulator of NK resistance.

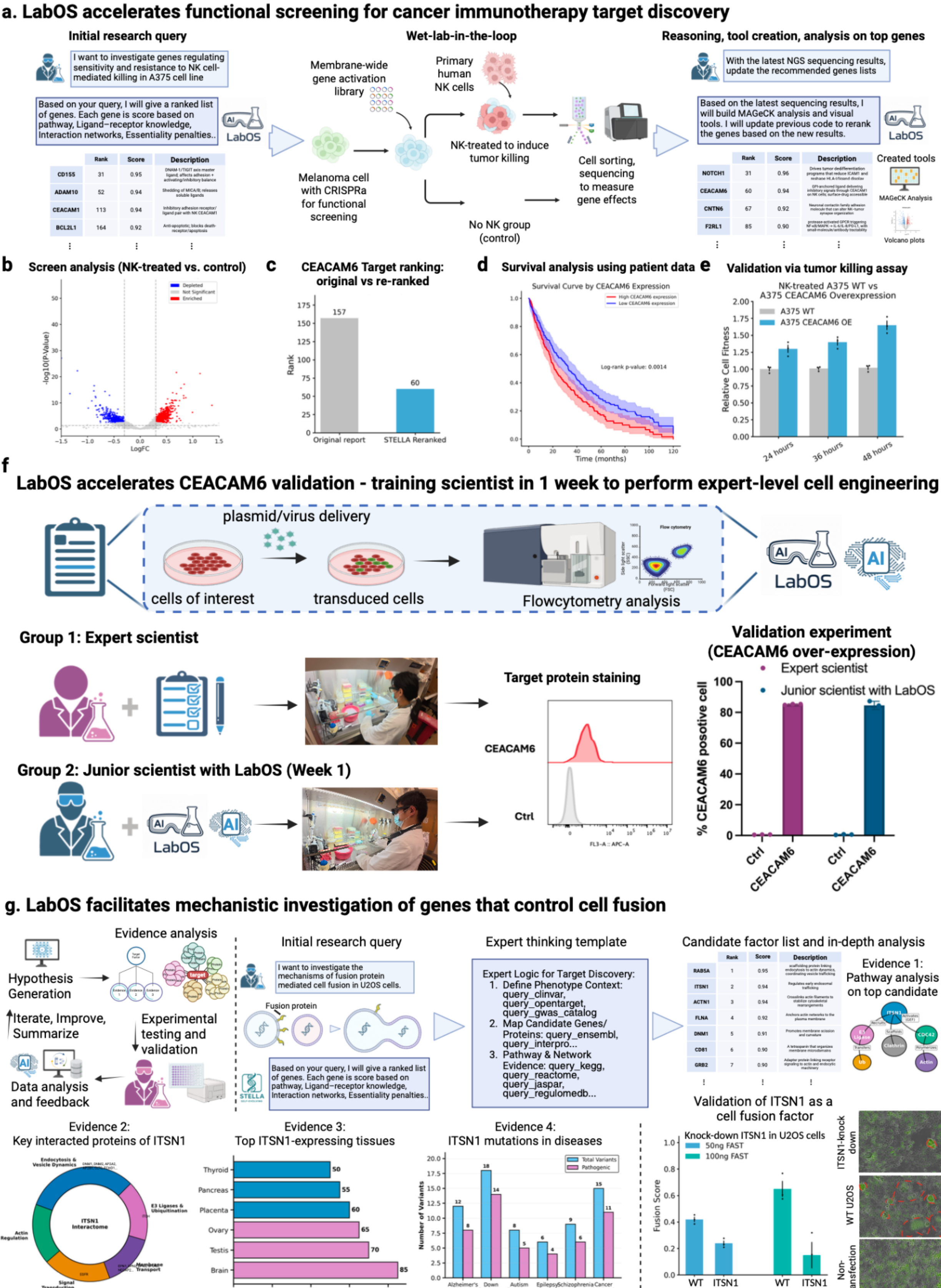

**Figure 4. LabOS applications in target discovery and mechanistic investigation. a-e, Functional screening study for Natural Killer (NK) immunotherapy target identification.** LabOS was prompted with the research task to identify regulators of tumor resistance to NK killing, based on results from a CRISPR activation screen in A375 melanoma cells co-cultured with primary human NK cells. **b-e,** AI-generated screen analysis reveals differential gene enrichment upon NK treatment (b), and the AI agent

re-ranked key target—*CEACAM6*—moving it from low to top-ranked position (c) and automated survival analysis using cancer patient data, stratified by CEACAM6 expression (d). Physical lab functional validation confirms enhanced NK resistance upon CRISPRa activation of CEACAM6 (e). **f,** LabOS accelerates CEACAM6 validation experiments by providing live copiloting to researchers for efficient cell engineering. The AI-XR agent can learn advanced skills in experiments and automatically document them, thus enabling junior scientists to perform complex multi-step workflow to over-express CEACAM6, enabling fast validation for cancer immunotherapy target discovery campaigns. **g**, LabOS applies to mechanistic investigation of cell fusion factors. LabOS proposes ITSN1 as a regulator of cell-cell fusion, providing its found evidence and reasoning trajectories. Experimental knock-out of top candidate gene ITSN1 confirmed this gene's impact on cell fusion in the physical lab, validating the AI co-scientist's hypothesis. **All photographs shown are of the authors.**

AI reasoning further generated explainable evidence by automatically performing survival analyses on *The Cancer Genome Atlas* (TCGA) datasets, stratified by CEACAM6 expression, thereby linking functional screening results to patient outcomes. Experimental validation using individual CRISPRa perturbation confirmed that CEACAM6 activation significantly increased tumor resistance to NK-cell killing (**Fig. 4a-f**).

### 5.2 LabOS copilots researchers in cell engineering experiments to accelerate CEACAM6 validation campaign

Reproducibility in advanced physical-lab domains is hindered by tacit expertise: critical steps are encoded in human memory and lab-specific "know-how," not necessarily in written protocols. For cell engineering, as an example, small deviations in the timing, reagent handling, or cell density and cell state assessment can drive large outcome variation, impeding transfer of skills across operators and laboratories.

For the NK immunotherapy target discovery campaign, a key experiment is overexpressing CEACAM6 in melanoma A375 cells and measuring NK-cell killing (**Fig. 4e**). This validation experiment is necessary to confirm CEACAM6's causal role in modulating NK resistance. We used the LabOS physical lab module to guide and automatically document these experiments, first from expert-level scientists (**Fig. 4f**). The LabOS system then provided context-aware guidance and training for junior scientists via the AI-XR glass. Here, LabOS acted as an AI tutor: recording expert practice, ambient data, digitizing key parameters, thereby enabling expert-level operation. The expression level of CEACAM6 protein in target cells confirmed that junior scientists successfully achieved over 80% efficiency. Overall, efficient gene delivery underpins construction of engineered cells for perturbation and drug screening. Our results demonstrated that the LabOS physical lab solution accelerated NK therapeutic target discovery through real-time, multimodal human-AI interactions.

### 5.3 LabOS guides mechanistic investigation of cell fusion regulator

Beyond functional screening, LabOS extends to mechanistic hypothesis generation and validation. In the next study, we prompt LabOS with the research task to identify key genes that control cell-cell fusion. Cell fusion is a fundamental biological process, key to basic biology (e.g. muscle development, viral infection all require cell fusion) and translational technology (e.g. cell fusion is a key step mediating efficient gene delivery).

LabOS proposed and ranked candidates using pathway enrichment, interaction priors, and functional evidence. The AI co-scientist prioritized *ITSN1* as a key regulator of cell-cell fusion, with automated evidence generation. Then human researchers used CRISPR interference (CRISPRi) coupled with cell fusion assay (induced by the fusogenic protein FAST [13]) in U2OS cells, to measure if ITSN1 perturbation will impact cell fusion phenotype. Researchers further performed quantitative imaging and cell-based assay and observed significant inhibition of cell fusion upon ITSN1 knockdown in the physical lab, confirming the role of ITSN1 (**Fig. 4g**).

### Summary

LabOS prototypes what an AI co-scientist can be: a system that sees, reasons, and helps run the digital and physical labs. By pairing AI agents with real-time, XR-guided human–AI-robotics interaction and data-driven reasoning, it enables faster discovery, reproducible training, and precise operation. Across use cases—hypothesis generation, automated documentation, error correction, rapid skill transfer, cell engineering experiment guidance, and insights into NK–tumor pathways and cell-fusion regulators—LabOS turns the lab into an adaptive, collaborative workspace where human scientists and AI work side-by-side to accelerate discovery, generate reproducible science, and improve together.

### Acknowledgement

We extend our sincere gratitude to the Nebius team for GPU server infrastructure support. We would like to acknowledge we procured the AR/XR glass hardware from Viture, and thank the Viture team for support on wireless streaming, Unity/Android app support. We would like to acknowledge we procured the VR/XR glass thanks to Dr. Ze Yuan from the UReality team and their software support. The results on cancer patient survival analysis here are in whole or part based upon data generated by the TCGA Research Network: https://www.cancer.gov/tcga. Mengdi Wang acknowledges support from by the Princeton AI Lab, Google Faculty Award, Office of Naval Research (ONR) under MURI grant N00014-24-1-2687.

### Author contributions

L.C. and M.W. conceived and supervised the project. L.C. led the research plan/organization, hardware/XR solution build, study design and paper writing. AI & Data: Z.Z, R.J. co-led the STELLA AI agents and benchmarking. X.W. led the data collection, annotation and human evaluation efforts, contributed to setting up XR glasses and live experimentation. Z.Z., Y.W., M.G. contributed to VLM training and benchmarking. D.S.,S.L., A.S. co-led 3D/4D modeling, object recognition and tracking of lab workflows. D.Y., X.W., P.L., Y.Z., S.S., N.Z. contributed to data collection, scientific annotation, and human evaluation of VLMs. Experiment: X. W and R.D. co-led NK experiments, to which J.W. also contributed and supervised. D.Y. led the fusogen experiment and co-lead stem cell experiments with X.W., J.C.W. co-supervised the stem cell experiments.

## Extended Data Figures

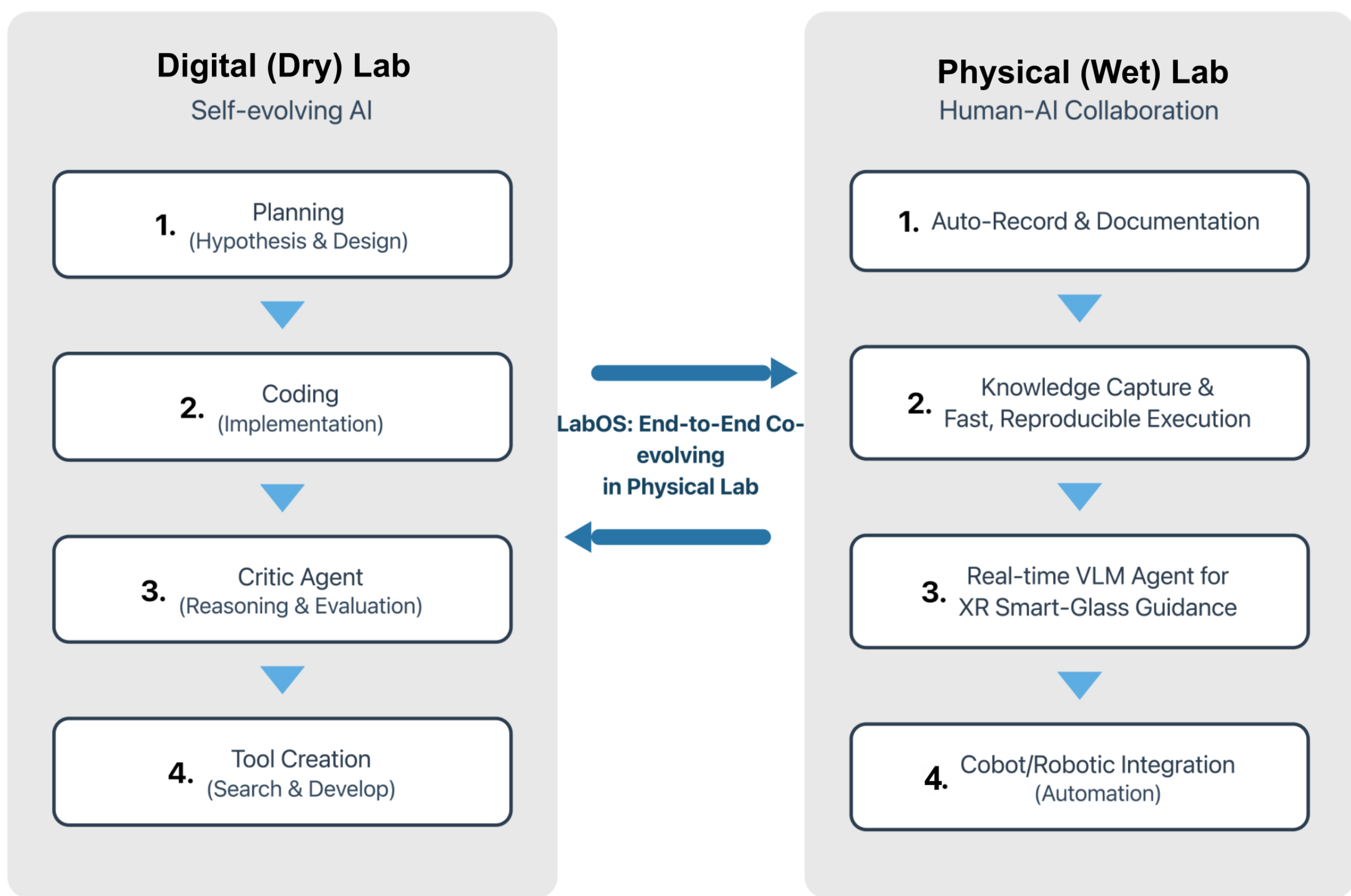

**Extended Data Figure 1. Simplified overview of the LabOS digital–physical co-scientist.** LabOS links a self-evolving AI “digital (dry) lab” with a human–AI collaboration “physical (wet) lab.” On the digital side, agents perform (1) planning (hypothesis and experiment design), (2) coding (implementation), (3) critique and reasoning, and (4) tool creation. On the physical side, LabOS supports (1) automatic recording and documentation, (2) knowledge capture and fast, reproducible execution, (3) a real-time vision–language agent for XR smart-glass guidance, and (4–5) cobot/robotic integration for automated execution. Bidirectional arrows indicate end-to-end co-evolution as digital agents learn from physical experiments, and thereby generates insights.

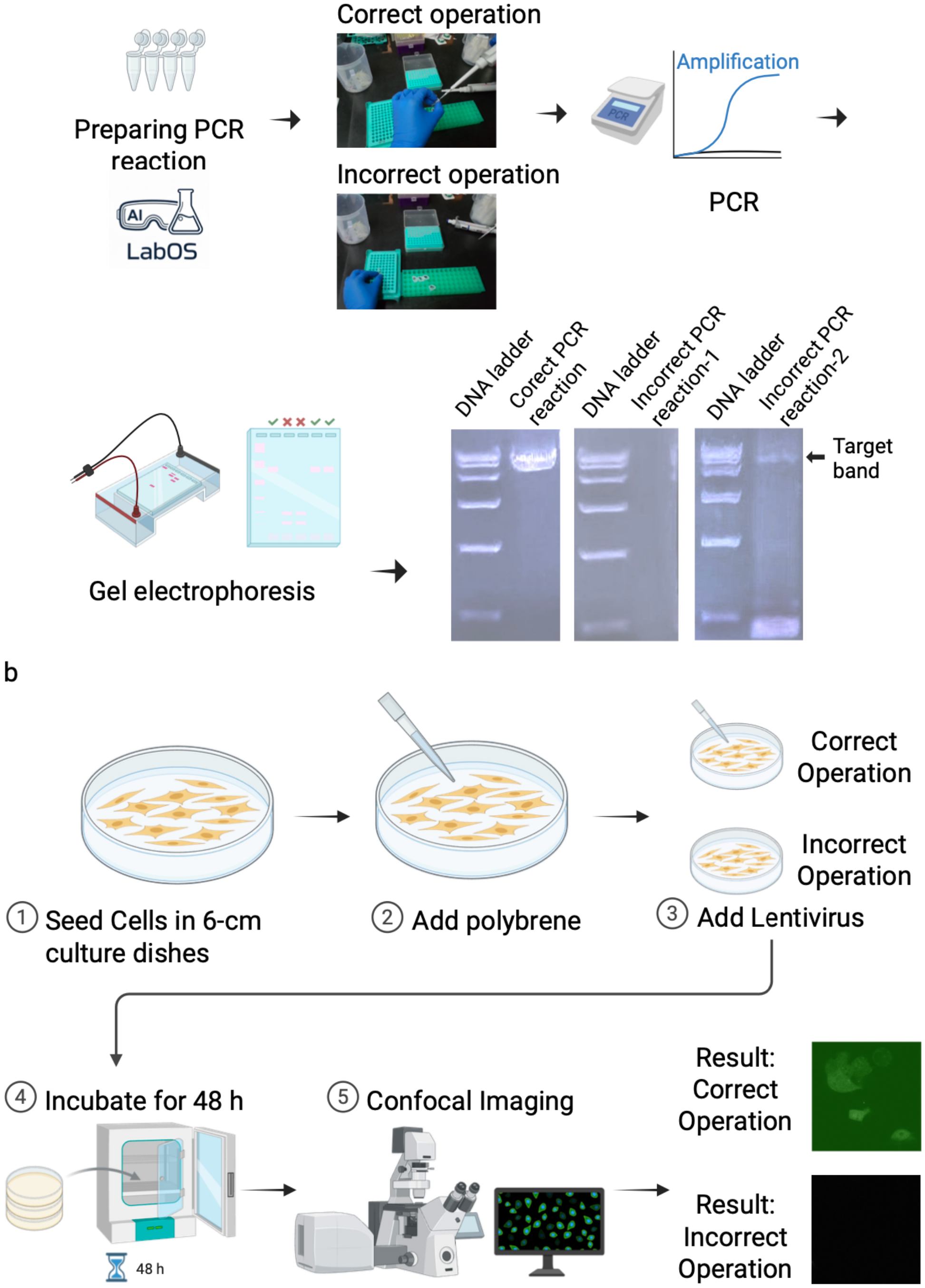

**Extended Data Figure 2. LabOS-guided error detection in common laboratory tasks.**
**a**, Example of a LabOS-guided PCR workflow. LabOS monitors preparation and key steps, distinguishing correct from incorrect operations during PCR setup and linking them to downstream outcomes on gel electrophoresis, where only correctly prepared reactions produce bright, expected PCR product band. **b**. Example of a LabOS-guided lentiviral transduction workflow. LabOS provides stepwise guidance for seeding cells, adding polybrene, adding lentivirus, incubating cells, and imaging. Correct execution yields robust fluorescent signal, whereas incorrect operation results in failed transduction (little fluorescence), illustrating how LabOS can help detect and correct errors in real time.

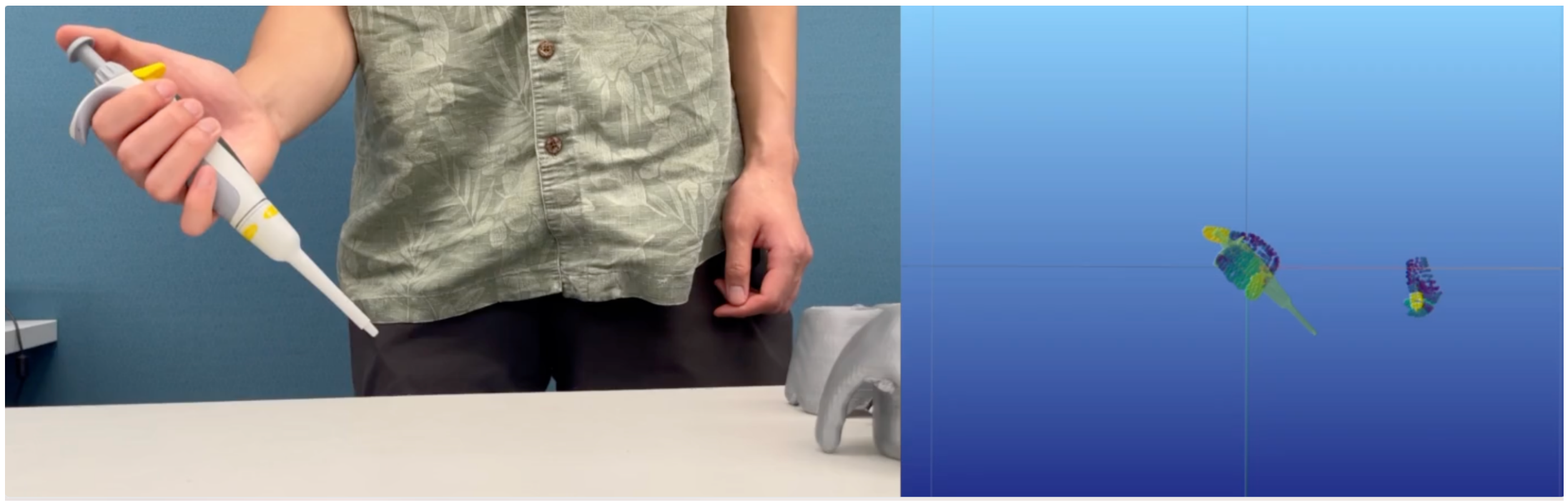

**Extended Data Figure 3. Tracking lab motion/operation with combined hand and object reconstruction.** LabOS uses hand-tracking and in-hand object pose–estimation models to jointly track a user's lab operation, as shown here, the scientist's hand and a pipette. Left, RGB view of a user manipulating a pipette. Right, reconstructed 3D trajectories and 6-DoF poses of the hand and pipette in a common coordinate frame after scale alignment, enabling precise modeling of hand–object motion during lab operations.